\documentclass[runningheads]{llncs}

\usepackage{eccv}

\usepackage{eccvabbrv}

\usepackage{graphicx}
\usepackage{booktabs}

\usepackage[accsupp]{axessibility}  

\usepackage[pagebackref,breaklinks,colorlinks,citecolor=eccvblue]{hyperref}

\usepackage{orcidlink}

\usepackage[marginal]{footmisc}

\begin{document}

\title{Knowledge NeRF: Few-shot Novel View Synthesis for Dynamic Articulated Objects} 

\titlerunning{Knowledge NeRF}

\author{Wenxiao Cai\inst{1} \and
Xinyue Lei\inst{1}* \and
Xinyu He\inst{1}* \and
Junming Leo Chen\inst{2} \and
Yangang Wang\inst{1}**
}


\authorrunning{W. Cai et al.}

\institute{Southeast University, Nanjing JS 211189, China\\
\email{\{wxcai, 213203641, 213201348, yangangwang\}@seu.edu.cn} \\
\and 
Northeastern University, Shenyang LN 110167, China\\
\email{leochenjm@gmail.com}
}

\maketitle
\footnotetext[1]{* Equal contribution}
\footnotetext[1]{** Corresponding author}

\begin{abstract}
We present Knowledge NeRF to synthesize novel views for dynamic scenes.
Reconstructing dynamic 3D scenes from few sparse views and rendering them from arbitrary perspectives is a challenging problem with applications in various domains. 
Previous dynamic NeRF methods learn the deformation of articulated objects from monocular videos. 
However, qualities of their reconstructed scenes are limited.
To clearly reconstruct dynamic scenes, we propose a new framework by considering two frames at a time.
We pretrain a NeRF model for an articulated object.
When articulated objects moves, Knowledge NeRF learns to generate novel views at the new state by incorporating past knowledge in the pretrained NeRF model with minimal observations in the present state. 
We propose a projection module to adapt NeRF for dynamic scenes, learning the correspondence between pretrained knowledge base and current states. 
Experimental results demonstrate the effectiveness of our method in reconstructing dynamic 3D scenes with 5 input images in one state. 
Knowledge NeRF is a new pipeline and promising solution for novel view synthesis in dynamic articulated objects.
The data and implementation are publicly available at \url{https://github.com/RussRobin/Knowledge_NeRF}. 
  \keywords{Novel View Synthesis \and Neural Radiance Fields \and Dynamic 3D Scenes \and Sparse View Synthesis \and Knowledge Integration}
\end{abstract}

\section{Introduction}
\label{sec_intro}

\begin{figure*}
	\begin{center}
  \includegraphics[width=1\textwidth]{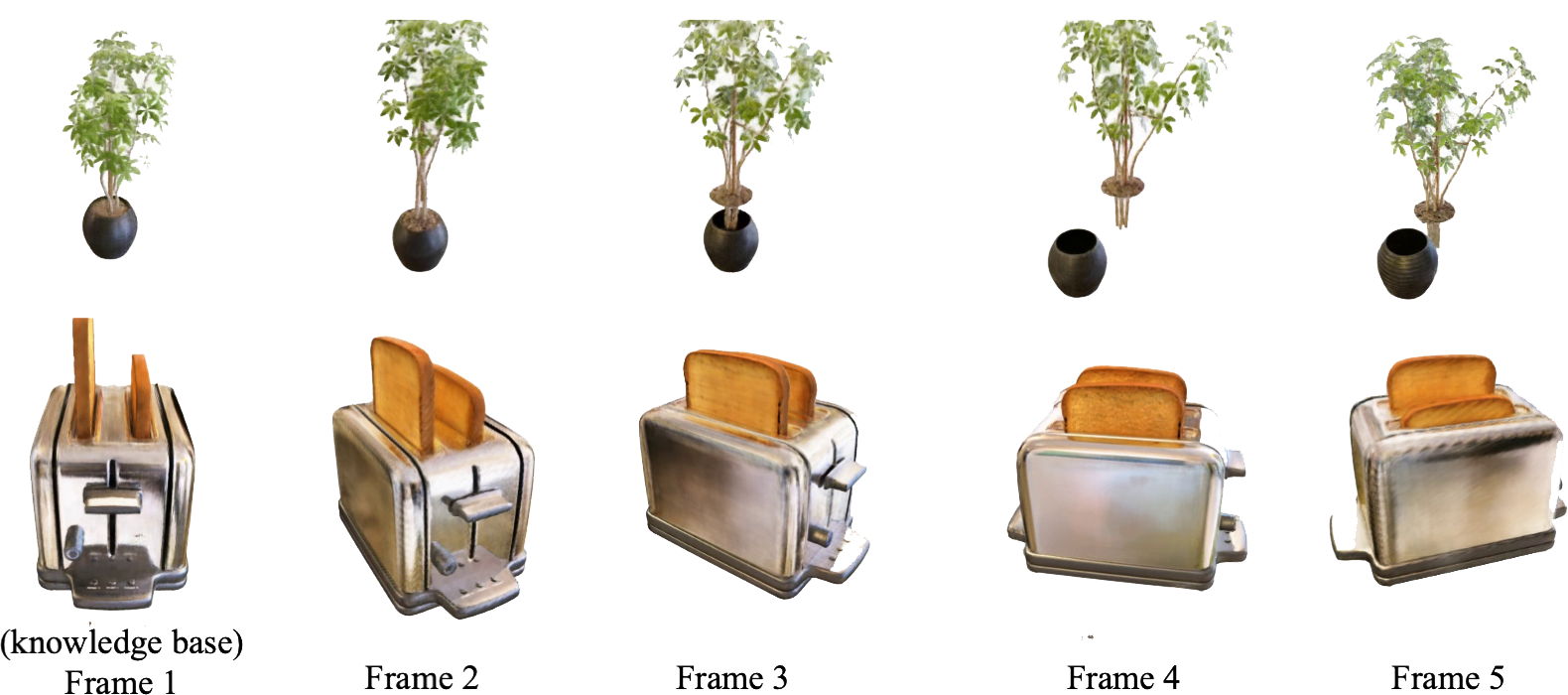}
  	\end{center}
  \caption{We propose a brand new framework, Knowledge NeRF, for dynamic novel view synthesis of articulated objects on. From a pretrained NeRF model in one frame, we train each following frame with 5 input images separately. The image quality of Knowledge NeRF is as clear as NeRF trained with abundant inputs.}
  \label{fig:teaser}
\end{figure*}

Reconstructing dynamic scenes from sparse views and dynamically rendering them from arbitrary new perspectives have long been a meaningful and challenging problem with wide applications in augmented reality~\cite{ar1,ar2}, virtual reality~\cite{vr1,vr2}, 3D content production~\cite{cp1,NeRFEditing}, sports~\cite{sport1}, hospitals~\cite{hospital1}, and the movie industry~\cite{movie1}. 
In dynamic scenarios, the movement of articulated objects is more worthy of attention as it's an essential foundation of robotic dynamics~\cite{articulate1}, VR/AR~\cite{articulate2,articulate3} and Human-Computer Interaction~\cite{articulate4,articulate5}.
While Neural Radiance Fields (NeRF)~\cite{NeRF} exhibits strong capabilities in representing 3D scenes, it lacks the ability to reconstruct scenes that change over time.
Dynamic NeRF methods~\cite{DNeRF,nerf-ds,nsff,nerfies} typically reconstruct three-dimensional scenes from a video, but the clarity of their reconstructed scenes is limited~\cite{DNeRF}. 
The challenge of dynamically reconstructing articulated objects clearly is yet to be addressed.

The input for dynamic NeRF problems~\cite{DNeRF} is a video capturing a moving object. Each state of the object corresponds to only one frame of the video, thus they cannot effectively learn the structure of the object itself, resulting in poor rendering quality. Their clarity is far lower than that of nerf trained with an adequate number of images at a single state.


To solve the obstacle, our key insight is to \textbf{infer the current appearance of the articulated object based on the knowledge from its past}.
We propose a novel framework to clearly reconstruct dynamic articulated objects by considering two frames at a time. 
For an articulated object undergoing rigid motion, we aim to reconstruct various states of the object in motion with as few images as possible. 
To achieve this, we first obtain a well-trained NeRF model of the object in a specific pose. As the object moves, we transfer knowledge from the known state to the current state, reconstructing the current 3D scene using only 5 images.
In contrast to previous methods that try to capture all states in the object movement at a time, we consider two states at a time: the pretrained knowledge base and the current state of interest.
Our goal is to achieve the same level of image clarity in rendering for articulated objects after movement as achieved by training NeRF directly with a large number of images.

Please consider such an application of Knowledge NeRF: you have a 3D model of an articulated object in your computer(we call it knowledge base). Obtaining the 3D model of its new state after movement doesn't require capturing nearly a hundred images and train NeRF again. Instead, only five images are needed. The new 3D model is easy to capture(only 5 images), fast to train(we only add a trainable light-weight module to a frozen NeRF), and as clear as the original model.

When an articulated object in our knowledge base undergoes motion, such as translation, rotation, scaling, changes in occlusion relationships, or the introduction of new parts, we employ a projection module to capture its motion. 
This module combines information from the pretrained knowledge base with minimal observations in the current scene. 
The projection module learns correspondence between the articulated object in the knowledge base and its current state, implicitly calculating relationships associated with object movements.
When querying the color and opacity information of a particular point in a new state, Knowledge NeRF utilizes the projection module to search the knowledge base for the corresponding point and then utilizes the information from that corresponding point.
The sample results of video sequences are shown in Fig. \ref{fig:teaser}, where we work on a pretrained frame 0 and infer with 5 images for each following frame independently.

The projection module is light-weight and integrated at the forefront of the basic NeRF framework. 
Through end-to-end training with mean-squared error (MSE) in conjunction with NeRF, the plug-and-play projection module learns the correspondence between points in the current state and the state within the knowledge repository.
The proposed Knowledge NeRF proves to be efficient and practical for articulated objects and common objects with deformation. 
Once a NeRF model is established for a specific state of a articulated object as knowledge base, the reconstruction of any state in the object's movement requires only 5 input images, and the rendered image is as clear as that in knowledge base. 
Through mathematical derivations and experiments on multiple datasets, this paper demonstrates the reliability of the proposed method.

In summary, we present Knowledge NeRF, a NeRF-based model for dynamically reconstructing articulated objects, considering 2 frames at a time. The main contributions of this paper are as follows:
\begin{itemize}
    \item We propose a new framework to dynamically synthesize novel views by considering two states of a articulated object at a time. The image quality rendered by our approach is as good as that achieved by directly training NeRF with abundant images.
    \item A light-weight projection module is proposed to incorporate information from the past knowledge base into the current state.
    \item We propose new real-world and synthetic datasets for evaluation of moving articulated objects.
\end{itemize}


\begin{figure}[t!]
	\begin{center}
    \includegraphics[width=0.85\linewidth]{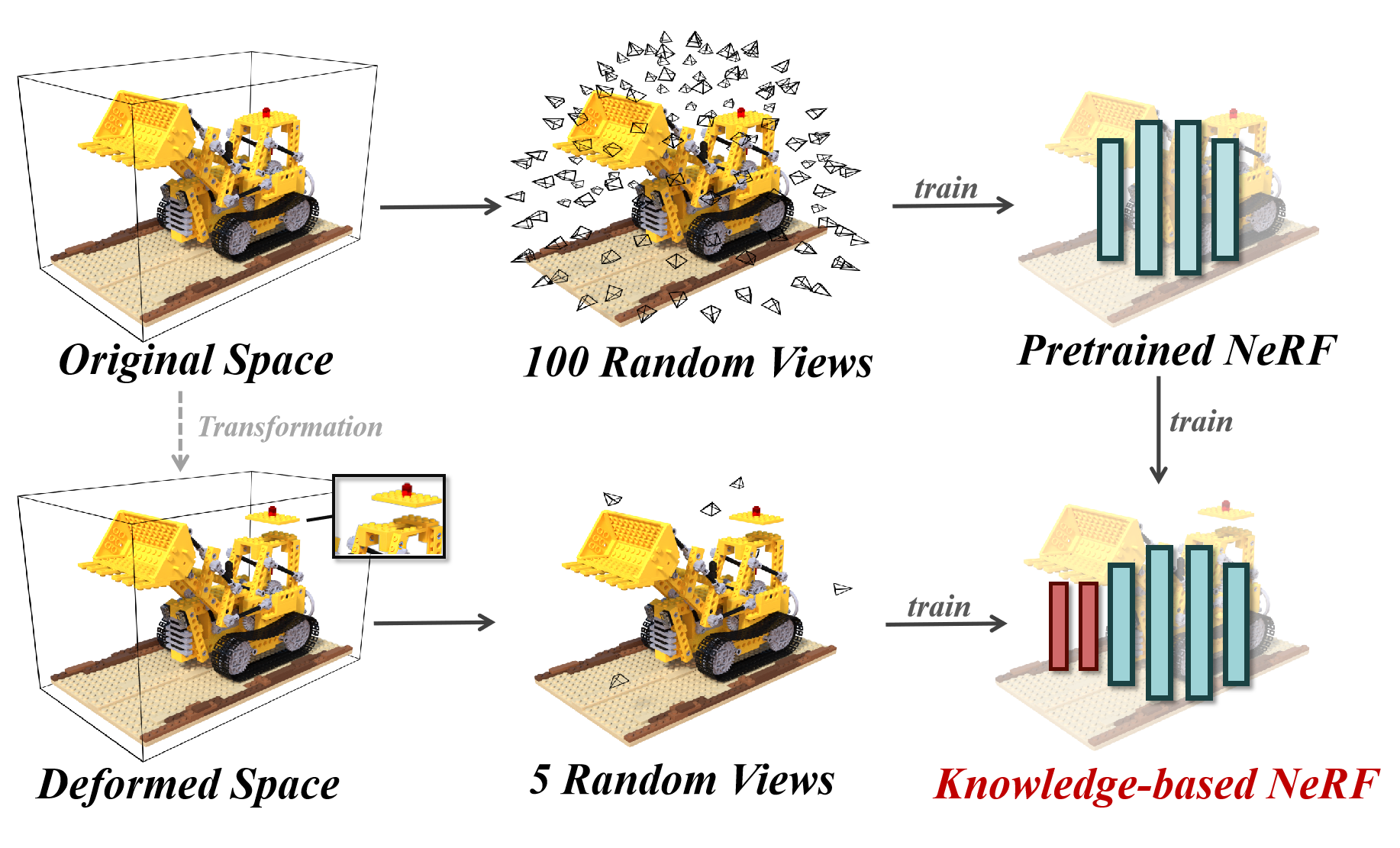}
	\end{center}
 	\caption{Overview of the proposed problem. Regardless of the articulated object undergoing any rigid transformation, we can train its 3D model in the new state using just five images.}
	\label{fig:overview}
\end{figure}

\section{Related work}
\label{sec_relatedwork}

\subsection{Dynamic NeRF}
HexPlane ~\cite{hexplane} and K-Planes~\cite{kplanes} explain the dynamic process as a projection problem of 4D volume, and apply an explicit method to project it onto six planes to simplify the calculation.HyperNeRF~\cite{HyperNeRF} borrowed the idea of DeepSDF~\cite{DeepSDF} to represent changes in scene topology by providing higher-dimensional inputs to the network. It can be used to reconstruct dynamic scenes with non-rigid objects and provides unprecedented accuracy. Similarly, D-NeRF~\cite{DNeRF} reconstructs and renders new views of objects with both rigid and non-rigid motion from a sparse collection of images that a single camera moves around a scene. However, considering the large amount of poses with the camera moving around, it still stays on the dense views in new views synthesis problems. 

Another approach, NeRF-Editing~\cite{NeRFEditing} allows users to make controlled shape changes to a NeRF of the scene and synthesize new views without retraining the network. Nevertheless, it requires other methods to extract the grid representation first. It also requires the manually selected objects to edit, and it cannot automatically segment different parts of the scene, nor can it handle large shape changes.
Unlike dynamic NeRF methods which learn from videos, we only consider two states in the articulated object movement. The new framework is able to generate novel views as clear as a well-trained NeRF model.

\subsection{Sparse View Reconstruction and Few-shot NeRF}
Numerous methods~\cite{IBRNet,pixelNeRF,SparseNeuS,NeuralRF,SRF,InfoNeRF} have successfully attempted sparse view reconstruction. RegNeRF~\cite{RegNeRF}, for instance, regularizes both scene geometry and appearance by utilizing rendered patches and implements a scene space annealing strategy. Depth prediction~\cite{depthprediction} recovered from different tasks frequently serves as a prior to aid sparse view reconstruction. DS-NeRF~\cite{DSNeRF} leverages depth information as supervision and claims to be able to reconstruct from as few as 2 or 5 views. Nevertheless, its view synthesis precision is limited to a large extent, and can not be improved with more inputs. Similarly, RGBD-NeRF~\cite{NeuralRF_PAMI} introduces depth from RGB-D images. The reconstructed network is pretrained using rendered images of the target and precise camera parameters, aiming to achieve more accurate results. Additionally, a small number of actually captured images are utilized for fine-tuning the network. Conversely, DietNeRF~\cite{DietNeRF} leverages semantics learned from CLIP~\cite{CLIP} before rendering new perspectives. This approach works well when the object can be depicted precisely in semantics. However, semantic priors also will introduce errors, leading to inaccuracies. While they endeavor to fit few-shot NeRF models by incorporating pretrained CLIP, we leverage pretrained NeRF models as our knowledge base.

In contrast to previous few-shot methods which primarily focus on reconstruction of static scenes, our approach aims to leverage past knowledge to accomplish dynamic 3D scenes in sparse views, namely 5 images as input. Compared with previous works, Knowledge NeRF is much more precise in dynamic scenes with sparse view inputs.

\section{Problem formulation}
\label{sec_formulation}
In this paper, our objective is to reconstruct the current scene $S_1$ using sparse view inputs, aiming to capture dynamic changes in the articulated object from a pretrained knowledge base $S_0$ to $S_1$. During this transformation, we observe two key phenomena: \\
$\bullet$ \textbf{Object surface of consistent points}. The surface of the articulated object is made up of the same points before and after deformation.
The focus of this paper is on the three-dimensional reconstruction of individual articulated objects, excluding considerations of their relationship with the scene background. \\
$\bullet$ \textbf{Particle opacity in NeRF training}.  In NeRF training for a single object, the particle opacity $\sigma$ tends to be either close to 0 or close to 1. Particles are categorized as either transparent air particles or opaque particles on the object's surface. \\

Based on these observations, we establish two fundamental assumptions:\\
$\bullet$ \textbf{Particle correspondence}. Particles on corresponding parts of the articulated object have a one-to-one correspondence from $S_0$ to $S_1$.\\
$\bullet$ \textbf{Air particle irrelevance}. Particles in the air may not have a correspondence, but their transparency makes them irrelevant in the reconstructed scenes.\\

Let $X_0$ denote the 3D location $(x_0,y_0,z_0)$ of a point in original scene $S_0$,and $X_1$ denotes the position $(x_1,y_1,z_1)$ of corresponding point in deformed scene $S_1$.
To mathematically represent the correspondence between particles $(x, y, z)$ from the current state to the previous state, we use the function $F_P$:
\begin{equation}
    X_0 = F_P(X_1) =
    \begin{cases}
        X_1, & \text{if } X_1 \text{ is in a static part}, \\
        T(X_1), & \text{if } X_1 \text{ is in a moving part}.
    \end{cases}
\end{equation}

Here, $T$ is a function representing rotation, translation, scaling and the introduction of new contents. 
$F_P$ is to approximate the movements of points, so it is designed to be lightweight. 
In this paper, the projection module is a made up of 4 fully connected layers, plugged before the frozen NeRF.
The input to $F_P$ is a three-dimensional position vector $X_1 = (x_1, y_1, z_1)$ in $S_1$, which is projected through $F_P(X)$ to the state $X_0 = (x_0, y_0, z_0)$ in $S_0$. 
For the viewing angles $(\theta, \phi)$, we assume $(\theta_1, \phi_1) \approx (\theta_0, \phi_0)$, since most points are not reflective and looks the same from all directions.

In summary, we place the $F_P$ module before the pretrained NeRF in the knowledge base, train $F_P$ to learn the projection relationship from $S_1$ to $S_0$, and then use the frozen NeRF in the knowledge base to compute the particle values $(r, g, b, \sigma)$ for the prompt $X_0 = (x_0, y_0, z_0, \theta_0, \phi_0)$. The function of the projection module $F_P$ is relatively simple, allowing for representation with a lightweight neural network. 
By combining past knowledge and utilizing a small amount of sparse view input, we can reconstruct the three-dimensional model of dynamic objects and generate new perspectives.
The pipeline is shown in Fig. \ref{fig:overview}.

\section{The proposed method}
\label{sec_method}
\subsection{Lightweight projection module}
To achieve the reconstruction of a three-dimensional model from extremely sparse viewpoints and accomplish novel view synthesis, we leverage the knowledge base of the model, which encapsulates its information at any given historical moment.
To facilitate the transfer of knowledge from the knowledge base to the current state, we introduce a lightweight projection module.
This module is designed to learn the correspondences between the object in its current state and knowledge base.
Drawing insights from our observations and appropriate simplifications in the problem formulation, we employ a projection module with Mean Squared Error (MSE) loss, utilizing a sparse set of view images. 
Through this module, the system learns how points on the object in the present state correspond to equivalent points in the knowledge base, enabling the rendering of the object in its current state using past knowledge. Simultaneously, the projection module is trained to disregard transparent points in the air.

\begin{figure*}[t!]
	\begin{center}
    \includegraphics[width=0.95\linewidth]{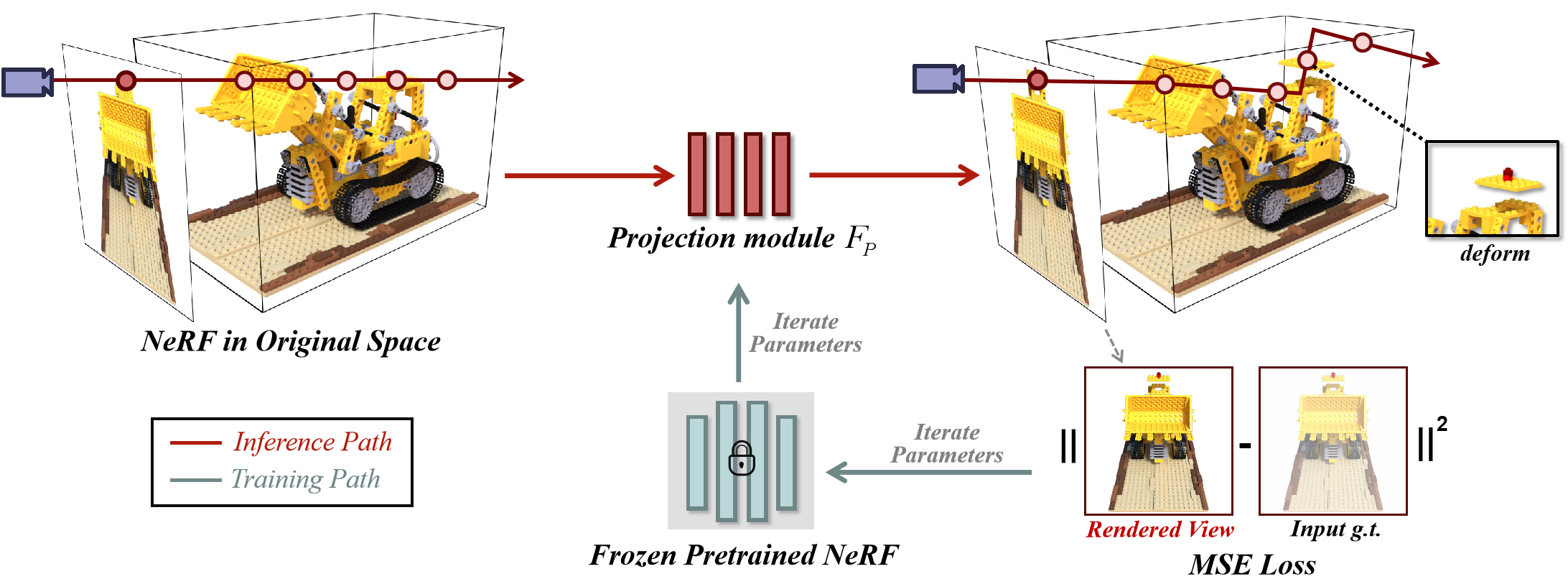}
	\end{center}
 	\caption{The proposed framework. A projection module is adopted to estimate the transformation between the original space and the deformed space. Within the deformed space, to render a specific pixel, we trace rays originating from the camera in the specified direction. To calculate the color and opacity of the points traversed by the rays, we utilize the projection module to find the corresponding point in the original space. Subsequently, a pre-trained NeRF model in the original space is utilized to ascertain the color and opacity of the identified point.}
	\label{fig:framework}
\end{figure*}

We first pretrain NeRF model $F_\Theta $ on the knowledge base $P_0$:
\begin{equation}
F_\Theta : (x_0, y_0, z_0, \theta_0, \phi_0) \rightarrow (r, g, b, \sigma).
\end{equation}
According to assumptions in Sec.~\ref{sec_formulation}, the particles should look the same from any viewing direction. Thus in $P_1$ and $P_0$, we approximate $(\theta_1, \phi_1) \approx (\theta_0, \phi_0)$.
The projection module on the small image set $P_1$ is trained to find:
\begin{equation}
F_P: (x_1, y_1, z_1) \rightarrow (x_0, y_0, z_0).
\end{equation}
In all, we try to find 4D output color and volume density by querying 5D input position and camera direction:
\begin{equation}
(r, g, b, \sigma) = F_\Theta(F_P(x_1, y_1, z_1), \theta_1, \phi_1).
\end{equation}

It's noteworthy that our method imposes no constraints on the spatial relationship between the positions where sparse view images are captured and those in the knowledge base. 
The reason behind this is quite simple: for a well-trained NeRF, we assume its outputs are close to ground truth in every view. Thus the input 5 views can find corresponding close-to-ground-truth views in the knowledge base to compare with.
We randomly sample abundant(typically 100) views to train original NeRF and then randomly sample 5 views to train Knowledge NeRF.
The proposed framework is shown in Fig. \ref{fig:framework}.


\subsection{Training pipeline}
To train efficiently, we initialize the projection module in Knowledge NeRF with $F_P(X_1) = X_0$ for any $X_1$. 
It is trained to learn  $X_0 = F_P(X_1) = X_1 + \Delta X$, where $X_0$ is transferred directly from input to output of the projection module.
Then we freeze NeRF and train projection with end-to-end loss, from MSE of rendered image and ground truth in sparse views.
Since some object segments would be occluded within the knowledge base but become visible in the new state, we employ a strategy of unfreezing all modules post-training and subsequently fine-tuning the system. This deliberate and sequential process ensures the adaptability of our model to dynamically evolving scenarios, allowing it to discern and incorporate previously hidden information in a refined manner.
Thus, the training pipeline of Knowledge NeRF consists of three parts:
\begin{enumerate}
    \item Train the NeRF model using images from $P_0$ in knowledge base.
    \item Initialize the projection module with $F_P(X_1) = X_0$ for any $X_1$, freeze the NeRF model, and end-to-end train the entire Knowledge NeRF model using a small amount of sparse view input from $P_1$.
    \item Unfreeze the NeRF model, and fine-tune using images from $P_1$, as there may be new occlusion relationships in $P_1$ compared to $P_0$. Visually, some points obscured in $S_0$ may appear in $P_1$.
\end{enumerate}

\subsection{Loss function}
In Knowledge NeRF, the positions of particles are projected from $P_1$ to $P_0$. They are then combined with camera directions to query color and volume density with $F_\Theta$ pretrained in $P_0$. The volume rendering adopted in NeRF~\cite{NeRF} is re-formulated as: 
\begin{equation}
\label{volume_render_eq}
    C(\textbf{r}) = \int_{t_n}^{t_f} T(t)\sigma[F_P(\textbf{r}(t))]\textbf{c}[F_P(\textbf{r}(t)),\textbf{d}] \, dt, 
\end{equation}
\begin{equation}
\label{volume_render_eq2}
    \mbox{where } T(t) = exp(-\int_{t_n}^{t_f} \sigma[F_P(\textbf{r}(s))] \, ds).
\end{equation}

In Eq.\ref{volume_render_eq} and Eq.\ref{volume_render_eq2}, $\textbf{r} = \overrightarrow{o} +t\overrightarrow{d}$ is the camera ray, $T(t)$ is accumulated transmittance alone the projected ray, from $t_n$ to $t_f$. $\sigma$ is the volume density of a particle, $\textbf{c}$ is its color, and $F_P(\textbf{r}(s))$ is the projected particle position. 


The proposed Knowledge NeRF tries to estimate the colors and volume density of chosen particles along a ray. As we adopt a coarse-to-fine structure, the training loss is the sum of squared error from true pixels for coarse and fine rendering:
\begin{equation}
    L_{mse} = \sum_{\textbf{r} \in R} [ \Vert \hat{C}_{c0}({F_P(\textbf{r}_1})) - C(\textbf{r}_1) \Vert^2 + \Vert \hat{C}_{f0}(F_P({\textbf{r}_1})) - C(\textbf{r}_1) \Vert^2 ],
\end{equation}
where $R$ is the set of rays in each batch, and $C(\textbf{r}_1)$ is the ground truth in training set $P_1$. $\hat{C}_{c0}$, and $\hat{C}_{f0}$ are coarse volume predicted, and fine volume predicted RGB colors, based on the $F_\Theta$ trained in knowledge base $P_0$. The inputs to $\hat{C}_{c0}$ and $\hat{C}_{f0}$ are projected rays $F_P(\textbf{r}_1)$.

\begin{figure*}[t!]
	\begin{center}
    \includegraphics[width=1\linewidth]{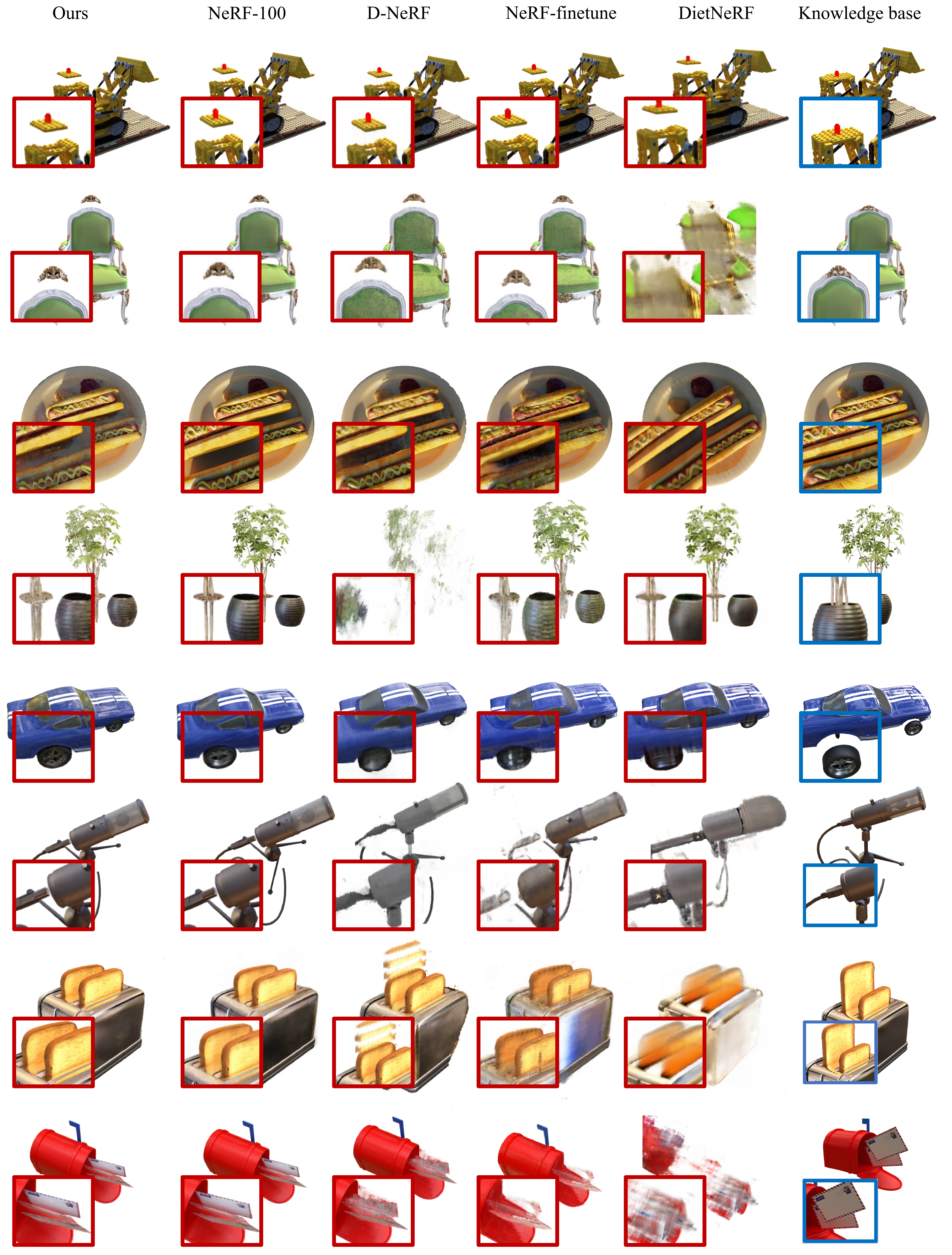}
	\end{center}
 	\caption{Qualitative results on 8 datasets. We compare sparse view reconstruction results of the proposed method. We also show the knowledge adopted in our method. Please note that views in the knowledge base and current state are randomly sampled.}
	\label{exp_fig}
\end{figure*}

\section{Experiments}
\label{sec_experiments}

\begin{table*}[h!]
\caption{
Quantitative Comparison of NeRF~\cite{NeRF}, DietNeRF\cite{DietNeRF} and the proposed Knowledge NeRF on synthetic datasets. NeRF-ft stands for finetuning NeRF in original space by 5 images in deformed space. We report their PSNR/SSIM (higher is better), and MSE/LPIPS/FID (lower is better). In NeRF-100 we report image qualities of NeRF trained with 100 images. Knowledge NeRF's image quality is comparable to NeRF-100.
}
\setlength{\tabcolsep}{1pt} 
\centering

\begin{tabular}{lcccccccccc}

\toprule
& \multicolumn{5}{c}{Lego} & \multicolumn{5}{c}{Chair}  \\
Method 
& MSE$\downarrow$ & PSNR$\uparrow$  & SSIM$\uparrow$ & LPIPS$\downarrow $ & FID$\downarrow $ 
& MSE$\downarrow$ & PSNR$\uparrow$  & SSIM$\uparrow$ & LPIPS$\downarrow$ & FID$\downarrow $ 
\\
\cmidrule(lr){1-1}  \cmidrule(lr){2-6} \cmidrule(lr){7-11} 
    DietNeRF
    & 7e-3 & 21.74 & 0.80 & 0.23 & 159.1
    & 7e-2 & 11.80 & 0.71 & 0.46 & 374.3 
    \\
    NeRF-ft
    & 3e-3 & 24.71 & 0.88 & 0.17 & 76.3 
    & 3e-3 & 27.17 & 0.90 & 0.13 & 222.6
    \\
    D-NeRF 
     & 4e-3 & 24.21 & 0.86 & 0.16 & 169.8  
    & 6e-3 & 20.58 & 0.85 & 0.25 & 290.6 
    \\
    Ours 
    & \textbf{1e-3} & \textbf{27.79} & \textbf{0.93} & \textbf{0.14} & \textbf{52.3} 
    & \textbf{1e-3} & \textbf{28.67} & \textbf{0.93} & \textbf{0.09} & \textbf{123.0}
    \\
\hline
    NeRF-100
    & 6e-4 & 31.86 & 0.96 & 0.11 & 50.6 
    & 1e-3 & 29.44 & 0.94 & 0.09 & 113.9
    \\
\toprule

& \multicolumn{5}{c}{Hotdog} & \multicolumn{5}{c}{Ficus} \\
Method 
& MSE$\downarrow$ & PSNR$\uparrow$  & SSIM$\uparrow$ & LPIPS$\downarrow $ & FID$\downarrow $ 
& MSE$\downarrow$ & PSNR$\uparrow$  & SSIM$\uparrow$ & LPIPS$\downarrow$ & FID$\downarrow $ 
\\
\cmidrule(lr){1-1}  \cmidrule(lr){2-6} \cmidrule(lr){7-11} 
DietNeRF
& 4e-3 & 23.72 & \textbf{0.88} & 0.22 & 100.8 
   & 7e-3 & 21.56 & 0.86 & 0.13 & \textbf{138.7}
   \\
    NeRF-ft
    & 5e-3 & 22.95 & 0.87 & 0.18 & 259.3
    & 6e-3 & 20.53 & 0.85 & 0.16 & 215.5
    \\
    D-NeRF 
     & 4e-3 & 23.21 & \textbf{0.88} & \textbf{0.16} & \textbf{87.48}
    & 6e-3 & 20.47 & 0.82 & 0.18 & 200.83
     \\
    Ours 
    & \textbf{3e-3} & \textbf{24.06} & \textbf{0.88} & 0.17 & 88.0
      & \textbf{5e-3} & \textbf{22.50} & \textbf{0.88} & \textbf{0.12} & 140.1 
    \\
\hline
    NeRF-100
    & 3e-4 & 34.58 & 0.96 & 0.08 & 68.3
     & 8e-4 & 30.64 & 0.97 & 0.04 & 84.9 \\
\toprule

& \multicolumn{5}{c}{Mic}  & \multicolumn{5}{c}{Car}  \\
Method 
& MSE$\downarrow$ & PSNR$\uparrow$  & SSIM$\uparrow$ & LPIPS$\downarrow $ & FID$\downarrow $
& MSE$\downarrow$ & PSNR$\uparrow$  & SSIM$\uparrow$ & LPIPS$\downarrow$ & FID$\downarrow $ 
\\
\cmidrule(lr){1-1}  \cmidrule(lr){2-6} \cmidrule(lr){7-11} 
    DietNeRF
    & 3e-3 & 15.70 & 0.83 & 0.27 & 218.1 
    & 6e-3 & 22.26 & 0.86 & 0.13 & 152.5 \\
    NeRF-ft
    & 2e-2 & 17.10 & 0.83 & 0.24 & 251.7
    & 6e-3 & 20.28 & 0.80 & 0.13 & 158.7 \\
    D-NeRF 
    & 4e-3 & 23.52 & 0.91 & 0.17 & 204.2
    & 5e-3 & 22.11 & 0.85 & \textbf{0.12} & 172.8 \\
    Ours
    & \textbf{7e-4} & \textbf{31.43} & \textbf{0.97} & \textbf{0.04} & \textbf{85.9}
    & \textbf{1e-3} & \textbf{22.15} & \textbf{0.86} & \textbf{0.12} & \textbf{124.1} 

    \\
\hline
    NeRF-100
    & 6e-4 & 32.15 & 0.97 & 0.03 & 85.0 
    & 1e-3 & 28.91 & 0.92 & 0.07 & 100.7\\
\toprule

& \multicolumn{5}{c}{Toaster} & \multicolumn{5}{c}{Postbox (Our dataset)}   \\
Method 
& MSE$\downarrow$ & PSNR$\uparrow$  & SSIM$\uparrow$ & LPIPS$\downarrow $ & FID$\downarrow $ 
& MSE$\downarrow$ & PSNR$\uparrow$  & SSIM$\uparrow$ & LPIPS$\downarrow$ & FID$\downarrow $
\\
\cmidrule(lr){1-1}  \cmidrule(lr){2-6} \cmidrule(lr){7-11} 
    DietNeRF
    & 2e-3 & 16.99 & 0.84 & 0.21 & 179.7  
    & 6e-3 & 12.21 & 0.79 & 0.31 & 187.5 
    \\
    NeRF-finetune
    & 3e-3 & 15.89 & 0.82 & 0.15 & 187.3
    & 3e-3 & 25.87 & 0.93 & 0.06 & 108.9
    \\
    D-NeRF  
     & 6e-3 & 19.60 & 0.85 & 0.15 & 97.8 
    & 6e-3 & 21.49 & 0.88 & 0.13 & 114.2 
    \\
    Ours 
    & \textbf{8e-2} & \textbf{20.48} & \textbf{0.88} & \textbf{0.12} & \textbf{87.2} 
    & \textbf{3e-4} & \textbf{34.01} & \textbf{0.97} & \textbf{0.04} & \textbf{58.0}
    \\
\hline
    NeRF-100
    & 3e-3 & 25.86 & 0.95 & 0.07 & 50.7
    & 1e-4 & 37.48 & 0.98 & 0.02 & 45.6
    \\
    
\bottomrule
\label{table-exp}
\end{tabular}
\vspace{-3em} 
\end{table*}

\subsection{Implementation details}
The NeRF model in the knowledge base follows the same architecture as the original NeRF paper, consisting of 8 layers used to encode volumetric density and color of the scene in canonical configuration. 
The projection module $F_P$ only needs to represent a simple spatial transformation function. Therefore, we use a 4-layer MLP to characterize it, with a fully connected layer for linear transformation at the output. 
For the details of the structure of the projection layer, please kindly refer to the Supplementary Materials.
In the second layer of the MLP and the final output layer, we apply two residual operations. 
We first encode the input position $X_1 = (x_1, y_1, z_1)$ into a 256-dimensional vector, and after applying $F_P$, obtain an output vector of the same size, $X_0 = (x_0, y_0, z_0)$. Concatenating $X_0$ with the position-encoded camera angles $(\theta_0, \phi_0)$, we input this into the NeRF to obtain color and density values $(r, g, b, \sigma)$ for volume rendering. 
Our loss, similar to NeRF, is the mean squared error of the rendered image colors. 
Consistent with NeRF, we use the coarse-to-fine structure with sampling points of $N_c = 64$ in the coarse network and $N_f = 128$ in the fine network in experiments. The model is trained with 800 $\times$  800 images during 150K iterations, with batch size 1024. As for the optimizer, we use Adam~\cite{Adam} with initial learning rate of 0.0005, $\beta = (0.9, 0.999)$ schedulers as $lr = base-lr * (0.1) ** (iter / 250000)$. 
We report that it takes about 2 hours on RTX 4060Ti (16G RAM) GPU to train the projection module.

\subsection{Datasets and Baseline methods}
We utilized the Blender dataset from NeRF~\cite{NeRF}, Shiny Blender dataset~\cite{shiny} and our proposed postbox dataset. We also provide real-world datasets(Telescope and Beats).
For each Blender dataset, we edit the articulated object with at least 2 states: knowledge base state and the current state. For each state, we randomly sample views to train models.
The modifications to articulated objects encompassed translations, rotations, scaling, alterations in occlusion relationships, and the introduction of new content. Details of the data utilized in the paper, along with corresponding usage specifications, are publicly accessible on our website.

We conducted a comparative analysis among Knowledge NeRF,  DS-NeRF~\cite{DSNeRF}, DietNeRF~\cite{DietNeRF}, and D-NeRF~\cite{DNeRF}. Additionally, we compared our approach with the original NeRF trained with abundant inputs (100 images).
Notably, DS-NeRF exhibited consistent failures across all scenarios in our experiments. 
DS-NeRF relies heavily on the relationship between objects and backgrounds for both the quantity and quality of the depth information in point clouds it employs as supervision. 
Consequently, in the main body of the paper, we exclusively present the comparison among Knowledge NeRF,  D-NeRF and DietNeRF. 
For NeRF-ft, please refer to the ablation studies section.
We train Knowledge NeRF with 5 images and D-NeRF with 75 images per state. In accordance with DietNeRF's paper, we train DietNeRF with 8 images, since it will perform reasonably in this case.
All views in our dataset are randomly sampled around the object.
It's noteworthy that Knowledge NeRF can perform as well as NeRF trained with 100 images, indicating good image quality. 
For further experimental details and a comparative analysis with DS-NeRF, please refer to our supplementary material.

\begin{table*}[h!]
\caption{
Quantitative Comparison on our proposed real-world dataset. Knowledge NeRF has comparable performance as NeRF trained with 100 images(NeRF-100).
}
\setlength{\tabcolsep}{1pt} 
\centering

\begin{tabular}{lcccccccccc}

\toprule
& \multicolumn{5}{c}{Telescope} & \multicolumn{5}{c}{Beats}  \\
Method 
& MSE$\downarrow$ & PSNR$\uparrow$  & SSIM$\uparrow$ & LPIPS$\downarrow $ & FID$\downarrow $ 
& MSE$\downarrow$ & PSNR$\uparrow$  & SSIM$\uparrow$ & LPIPS$\downarrow$ & FID$\downarrow $ 
\\
\cmidrule(lr){1-1}  \cmidrule(lr){2-6} \cmidrule(lr){7-11} 
    DietNeRF
    & 2e-2 & 15.83 & 0.70 & 0.48 & 18.44
    & 3e-2 & 14.28 & 0.60 & 0.67  & 56.82
    \\
    NeRF-ft
    & 0.16 & 7.92 & 0.54 & 0.60 & 23.71
    & 3e-3 & 24.27 & 0.77 & 0.44 & 36.65
    \\
    D-NeRF 
    & 9e-3 & 20.31 & 0.82 & 0.21 & 12.52
    & 4e-3 & 23.24 & 0.73 & 0.52 & 38.91
    \\
    Ours 
    & \textbf{1e-3} & \textbf{29.67} & \textbf{0.95} & \textbf{0.10} & \textbf{5.87} 
    & \textbf{1e-3} & \textbf{27.36} & \textbf{0.79} & \textbf{0.41} & \textbf{25.19}
    \\
\hline
    NeRF-100
    & 1e-3 & 29.29 & 0.95 & 0.09 & 5.38 
    & 3e-3 & 24.45 & 0.77 & 0.42 & 26.50
    \\
    
\bottomrule
\label{table-exp-realworld}
\end{tabular}
\vspace{-3em} 
\end{table*}

\begin{figure*}[ht!]
	\begin{center}
    \includegraphics[width=1\linewidth]{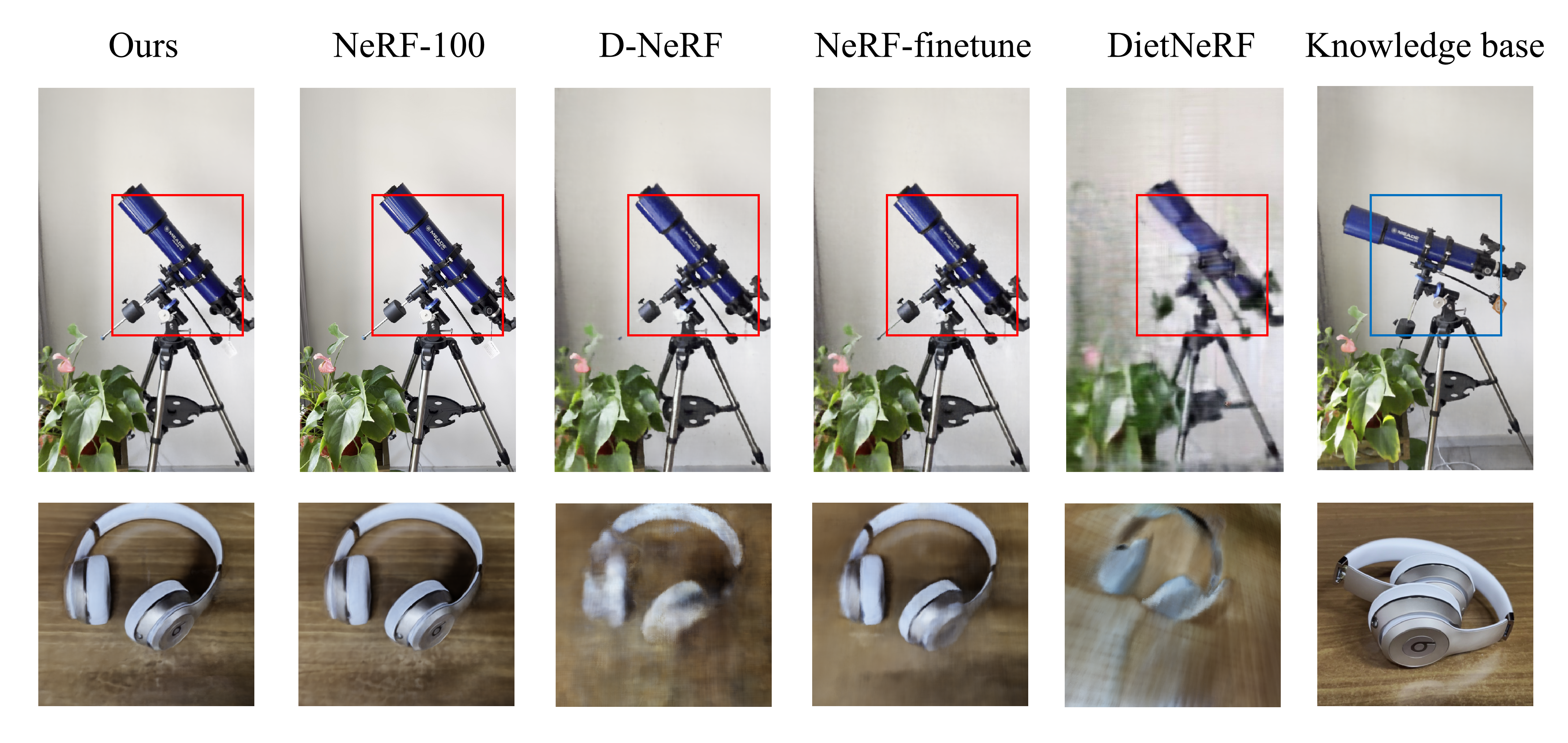}
	\end{center}
 	\caption{Quanlitative results on our proposed real-world dataset.}
	\label{fig_realworld}
\end{figure*}

\subsection{Qualitative comparison}
In Blender, we perform \textbf{translations, rotations, scaling and the introduction of new contents} on a subset of an articulated model. Our proposed approach, Knowledge NeRF, excels at sparse view reconstruction tasks with just 5 images, as it can effectively learn the movements and corresponding relationships between the new state and the knowledge base. NeRF, on the other hand, struggles to generate novel views from such a limited number of images. With some improvement in performance at 10 images, DietNeRF~\cite{DietNeRF} leverages higher-dimensional semantic information but still falls short compared to Knowledge NeRF using only 5 images. DietNeRF performs poorly with only 5 images. 
As reported by its authors, DietNeRF would perform reasonably with 8 images. However, in our experiments, DietNeRF still can't be as clear as the proposed Knowledge NeRF even trained with 8 images.
Please refer to Fig.~\ref{exp_fig} for the results on synthetic datasets, and Fig.~\ref{fig_realworld} for results on our proposed real-world dataset.

\begin{figure*}[ht!]
	\begin{center}
    \includegraphics[width=1\linewidth]{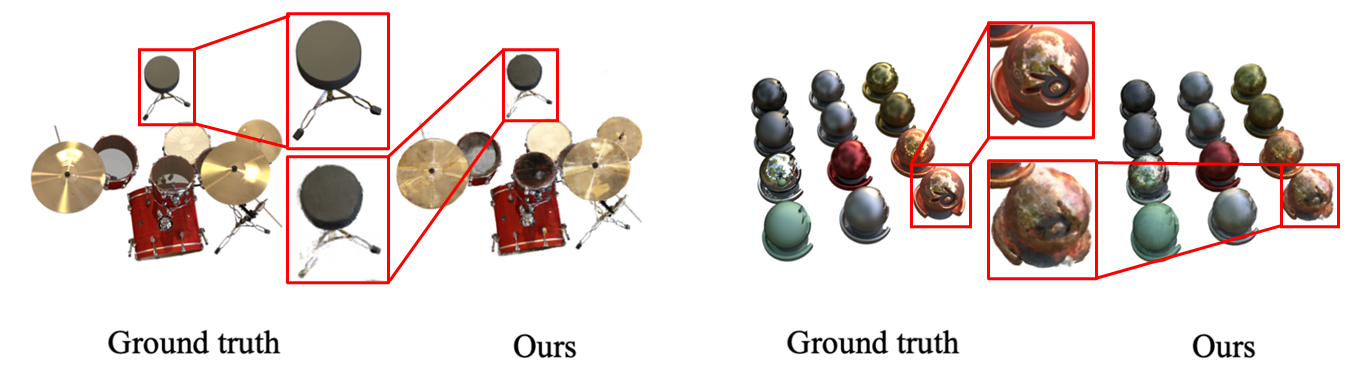}
	\end{center}
 	\caption{Failure cases on 2 datasets. Knowledge NeRF fails because of reflection. We move the chair and rotate the ball, and show their details in comparison.}
	\label{exp_fail}
\end{figure*}

\textbf{Occlusion}. 
In the Ficus dataset, the plant roots are initially occluded and emerge in the new state. In the frozen training results, the plant roots appear blurry because the projection module fails to find a corresponding part in the knowledge base. However, upon thawing, the roots become clear. In the Hotdog dataset, the middle portion between two hotdogs is obstructed in the knowledge base due to shadows cast by the light source, resulting in a black shadow. In the new state, the two hotdogs are moved, revealing a clear middle portion. Our projection module understands such spatial transformations and successfully reconstructs the clear middle portion during the thawed training process. We posit that for a smaller portion, five sparse view images are sufficient to train the novel view synthesis model. In the Hotdog dataset, the middle portion between two hotdogs is obscured.

\textbf{New content}.
In the Postbox dataset, the new state includes an additional envelope identical to one in the knowledge base. Our Projection module successfully recognizes that the newly added envelope originates from the knowledge base. If we denote the extra envelope and the original envelope as $e_1$ and $e_2$, respectively, this implies that $F_P(e_1) = F_P(e_2)$.

\subsection{Quantitative comparison} 
\textbf{Metrics}.
The  Mean Squared Error (MSE), Peak Signal-to-Noise Ratio (PSNR), Structural Similarity (SSIM)~\cite{ssim} and Learned Perceptual Image Patch Similarity (LPIPS)~\cite{lpips}, Fréchet Inception Distance (FID)~\cite{fid}, and  Kernel Inception Distance (KID) ~\cite{kid1, kid2} are rated in Table~\ref{table-exp}. 
 MSE, PSNR and SSIM measure the similarity between the model outputs and ground truth. LPIPS measures the perceptual image quality.  FID and KID are commonly used in generative models to measure sample quality.

Statistical results on synthetic datasets are reported in Table~\ref{table-exp}, while the real-world dataset is reported in Table~\ref{table-exp-realworld}. The image quality of Knowledge NeRF is always comparable to NeRF trained with 100 images (NeRF-100).

\subsection{Ablation studies}
To investigate the effectiveness of the proposed projection module, we conducted experiments involving NeRF fine-tuning. Building upon a pretrained NeRF in the knowledge base, we utilized five images of articulated objects in new states and employed MSE loss for NeRF training. The results obtained from the knowledge NeRF outperformed those from the fine-tuned NeRF. The experimental findings are reported in Table~\ref{table-exp}.

\begin{figure}[t]
	\begin{center}
    \includegraphics[width=1\linewidth]{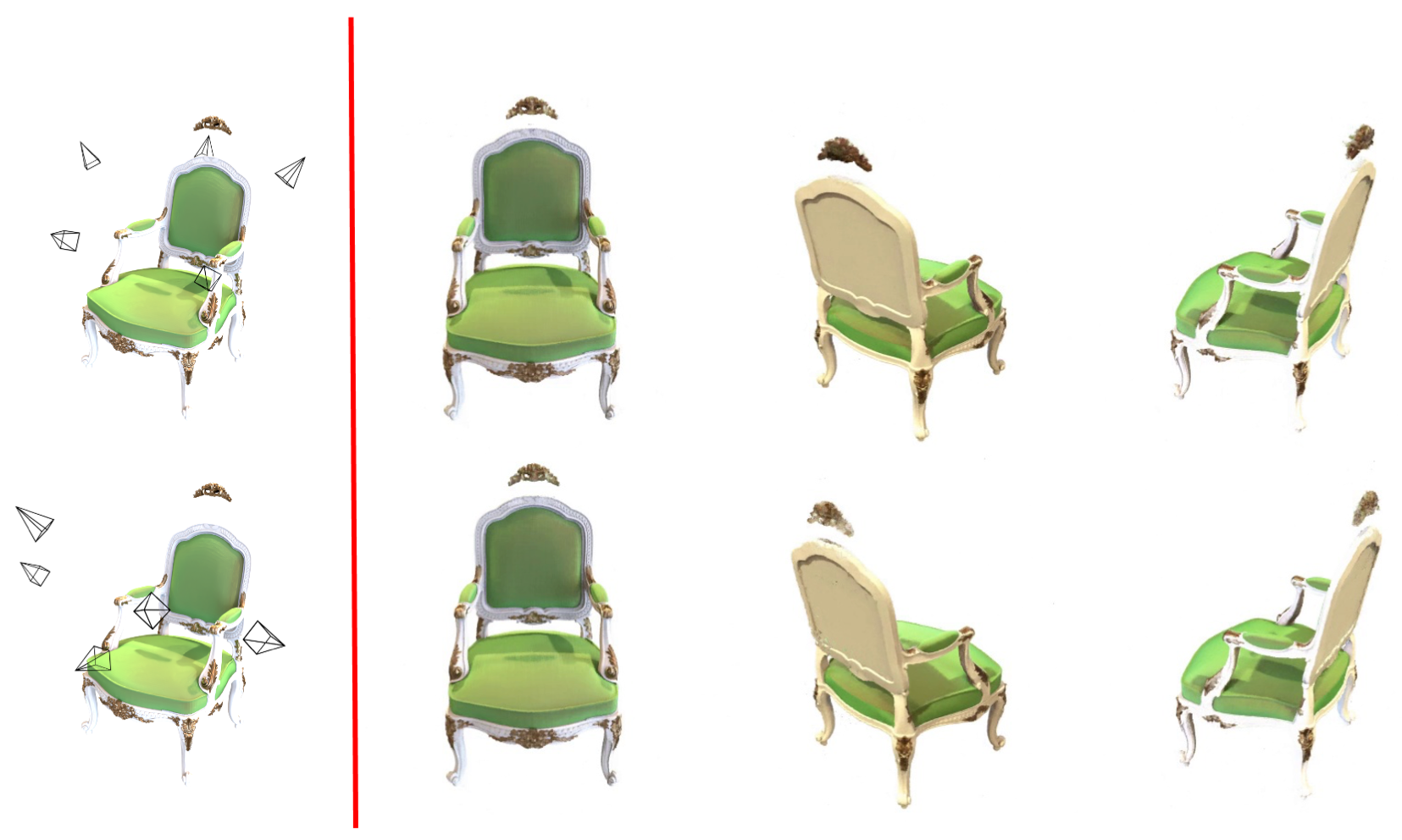}
	\end{center}
 	\caption{Left: input views in current states. Right: rendering results from 3 views. Up: random samples around the chair. Down: only sample in front of the chair.}
	\label{fig:supp-ablation}
\end{figure}

We also study how the distribution of the camera poses impacts Knowledge NeRF. In one experiment, we randomly sample images and achieve good reconstruction results. In another experiment, in the current state, our view is only in front of the object, with no input view behind the object. However, we obtain similarly clear reconstruction results. Therefore, our method is insensitive to the camera position. Please refer to Fig.~\ref{fig:supp-ablation}.

\subsection{Failure cases}
In accordance with the fundamental assumptions and derivations previously elucidated, it is imperative to acknowledge that our methodology is not conducive for application in scenarios characterized by reflection. In datasets incorporating reflections, instances arise wherein certain regions, originally intended to remain unoccupied, exhibit partial manifestations of objects, as shown in Fig.~\ref{exp_fail}. This phenomenon can be attributed to the limitations imposed by reflection, impeding the accurate estimation by our projection module of the spatial correspondences between the current state of objects and their representations within the knowledge base, as well as the relationships governing object movements.

\section{Discussions}

\textbf{Limitations}
The proposed Knowledge NeRF excels in accomplishing novel view synthesis with few inputs. However, in the case of dynamic articulated objects, each distinct motion or action may necessitate retraining the projection module. This requirement introduces a substantial overhead in terms of time and computational resources. The process of retraining for every action can be impractical, particularly in real-time applications where efficiency is paramount.
Addressing these limitations poses opportunities for future research. Strategies to enhance model generalization and reduce the need for frequent retraining could significantly improve the scalability and applicability of Knowledge NeRF in dynamic scenes.

\textbf{Broader impacts and future works}.
It’s important to acknowledge that our paper doesn’t explicitly discuss broader impacts in the proposed Knowledge NeRF, such as fairness or bias. Further research into how our novel view synthesis and NeRF models may interact with other aspects of 3D reconstruction and image processing is encouraged.
The pipeline of Knowledge NeRF may integrate 3D reconstruction with pose estimation in the future. Firstly, it estimates the poses from the knowledge base, then, using a small number of images, it estimates the pose changes of objects and infers new 3D models.

\section{Conclusions}
\label{sec_conclusions}
In this paper, we present Knowledge NeRF, a novel approach for addressing the challenge of sparse-view reconstruction and few-shot novel view synthesis in the context of articulated objects and other common objects with simple movements. By integrating past knowledge within the NeRF framework and employing a lightweight projection module, our method efficiently reconstructs three-dimensional scenes from a small number of sparse views.
Experimental results showcase the effectiveness of Knowledge NeRF in capturing dynamic changes in object states. Leveraging a pretrained NeRF as knowledge significantly reduces the data required for reconstructing new states, making our approach practical for scenarios with limited input.
Knowledge NeRF can precisely generate novel views with only 5 images as input.
The proposed method opens avenues for efficient sparse-view reconstruction, offering a valuable contribution to the field of 3D scene representation. 
We believe our approach has the potential to impact various applications, including augmented reality, virtual reality, and 3D content production, paving the way for advancements in dynamic scene understanding.

\clearpage  

%
%
\bibliographystyle{splncs04}
\bibliography{main}
\end{document}